\documentclass{ifacconf}

\usepackage{graphicx}      
\usepackage{natbib}        

\usepackage{amsmath} 
\usepackage{amssymb} 
\usepackage{booktabs}
\usepackage[justification=centering]{caption}
\usepackage{graphicx}
\usepackage{float} 
\usepackage{caption}
\usepackage{subcaption}

\usepackage[ruled, vlined, linesnumbered]{algorithm2e}

\usepackage{xcolor}
\usepackage{soul}
\usepackage[switch]{lineno}

\xdefinecolor{redZ}{RGB}{234,29,93}
\xdefinecolor{blueZ}{RGB}{19,106,213}
\xdefinecolor{yellowZ}{RGB}{251,138,46}
\xdefinecolor{greenYL}{RGB}{0, 191, 0}

\begin{document}
\begin{frontmatter}

\title{
Residual Reinforcement Learning for Robot Teleoperation under Stochastic Delays
}
\author {Kaize Deng}, 
\author {Zewen Yang\thanksref{footnoteinfo}} 

\thanks[footnoteinfo]{Corresponding author (e-mail: zewen.yang@tum.de).}
\address {Technical University of Munich, 80333 Munich, Germany \\(e-mail: kaize.deng@tum.de, zewenyang@tum.de).}

\begin{abstract} 
Stochastic communication delays in teleoperation introduce signal discontinuities that undermine control stability and degrade control performance. Consequently, the conventional reinforcement learning (RL) methods struggle with the delayed observations due to the delay-induced observations, leading to high-frequency chattering. To address this, we propose a hybrid control framework, delay-resilient RL, integrating a state estimator utilizing Long Short-Term Memory (LSTM) with a residual RL policy, which is resilient to stochastic delays. The LSTM reconstructs smooth, continuous state estimates from delayed observations, enabling the RL agent to learn a residual torque compensation policy that balances tracking accuracy with velocity smoothness. Experimental validation on Franka Panda robots demonstrates that our approach significantly outperforms the state-of-the-art baselines, ensuring robust and stable teleoperation even under high-variance stochastic delays.
\end{abstract}


\end{frontmatter}
\section{Introduction}
Teleoperation, a critical technology in modern robotics, enables human operators to control remote manipulators across geographical distances via networked communication. This capability facilitates expert intervention in environments where direct physical presence is impractical or hazardous, spurring transformative applications in domains such as remote surgery \citep{Choi_Cureus2018_Telesurgery}, distributed manufacturing \citep{Manupati2017_Telefacturing}, and even space robotic coordination \citep{Ruoff_1994_TeleoperationBook}. 

\begin{figure}[t]
    \centering
    \includegraphics[width=0.8\linewidth]{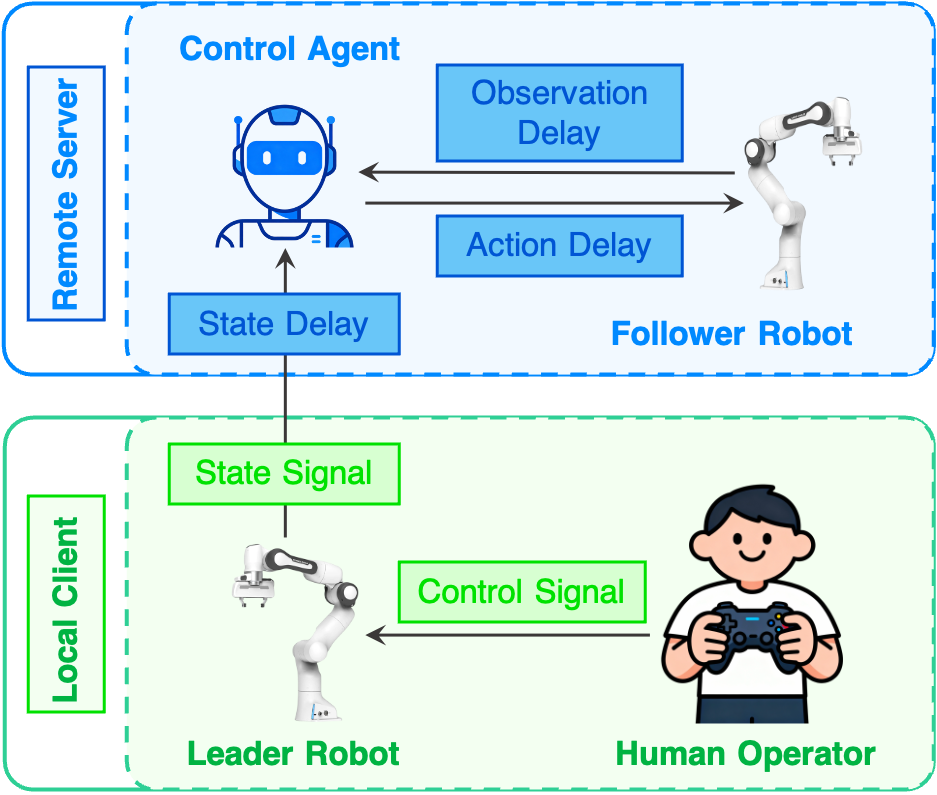}
    \caption{Teleoperation architecture with stochastic communication delays. The time-varying state delay $\omega_s^t$ is induced by client-server transmission latency; the action delay $\omega_a^t$ and observation delay $\omega_o^t$ arise from the bidirectional delays within the agent-follower control loop.
    }
    \label{fig_Teleoperation_architecture}
\end{figure}

Architecturally, a teleoperation system comprises two primary components: a local leader device, which captures the human operator's commands, and a remote follower manipulator, which is tasked with executing the corresponding movements in the distant environment (see Fig.~\ref{fig_Teleoperation_architecture}). 
First, the networked communication channel between the local and remote sites introduces a stochastic, time-varying state transmission delay. Specifically, the state signal of the leader is sent over the network to the remote reinforcement learning (RL) agent and arrives with a delay $\omega_s^t \in \mathbb{R}_{\ge 0}$ at time $t\in \mathbb{R}_{>0}$. 
This information gap fundamentally transforms the problem from a fully observable Markov Decision Process (MDP) into a Partially Observable MDP (POMDP), in which an optimal action cannot be computed directly from the past states.
Second, the agent–follower link is subject to inherent bidirectional delays. The observation delay $\omega_o^t \in \mathbb{R}_{\ge 0}$ affects the sensor feedback from the follower and directly induces delay observability. 
Therefore, at any time $t$, the controller has access only to delayed state information corresponding to $t - \omega_o^t$, rather than the true current state of the follower robot. 
Third, stochastic variability in the action delay $\omega_a^t \in \mathbb{R}_{\ge 0}$, commonly referred to as jitter, disrupts the command stream along the same agent–follower link. 
All these factors are particularly detrimental to low-level Proportional–Derivative (PD) control laws, i.e., the asynchronous arrival of commands injects artificial variability into the derivative of the tracking error, thereby inducing high-frequency oscillations, mechanical vibration, and chattering in the closed-loop system.

\subsection{Related Work}
Prior work has addressed the challenges induced by communication delays in teleoperation through several paradigms. 
Passivity-based control frameworks leverage energy conservation principles to establish formal stability guarantees~\citep{Niemeyer_Slotine_1991_StableAdaptiveTeleoperation, Mujcic_InternetTeleoperation_2019}. 
By mapping impedance signals into scattering variables, these methods ensure passivity of the communication channel and, by extension, closed-loop stability irrespective of constant delay magnitude. 
However, the theoretical stability is not guaranteed when delays are stochastic and time-varying rather than constant in real-world networks. Moreover, the requisite signal transformation compromises system transparency, leading to degraded tracking fidelity and attenuated force reflection. 
Model-based predictive strategies, such as the Smith Predictor \citep{Smith_1957}, attempt to explicitly compensate for latency by simulating the system response forward in time.
However, this approach is predicated on accurate a priori knowledge of both plant dynamics and delay characteristics. In scenarios where delays exhibit stochastic variation or significant model–plant mismatch exists, predictive performance deteriorates substantially.

More recent state-of-the-art (SOTA) methods have leveraged RL to tackle these uncertainties. 
Bypassing the need for such precise, model-based prediction, one category of work uses RL to autonomously tune controller gains \citep{Wang2007_AdaptivePIDRL, Lee_OceanEng2020_RLAdaptivePIDDPS, Zhang2021_RLDelayControl,huang2026contactsafereinforcementlearningpromp}. These methods, however, typically model the problem as a MDP, which fundamentally assumes the agent can observe the complete, current state to make an optimal decision. 
This core assumption is violated by observation delays, where the agent receives only outdated states. 
Policies trained under this false MDP assumption are thus acting on past information, leading to suboptimal and potentially unstable decisions \citep{Huang2019_RLTimeDelay, Katsikopoulos_TAC2003_MDPDelays}.

To address this POMDP, a common strategy is state augmentation \citep{Nath_ACM2021_RevisitingStateAugmentation}, which stacks past observations and actions into an expanded state representation. However, as delay length grows, the state space expands dramatically, invoking the curse of dimensionality and making learning computationally intractable. Moreover, this fixed-history approach often fails to generalize across variable or highly stochastic delay patterns.
Another methodology is model-based Reinforcement Learning (MBRL). 
The core idea is to learn a predictive model of the environment's 
dynamics to compensate for latency \citep{Barde2020_HumanSpeed}. 
A naive application, Action-Buffer based State Prediction (ABSP), 
is computationally prohibitive as it requires re-simulating the 
entire future trajectory at every time step. To address this 
inefficiency, the Predictive Model Delay Correction (PMDC) framework 
\citep{McCutcheon_IROS2023_Adaptive} introduced the more efficient 
State-Buffer based State Prediction (SBSP). SBSP maintains a buffer 
of predicted future states and updates it incrementally, making it 
a computationally viable solution for real-time control.
However, despite this computational advance, the SOTA PMDC framework 
exhibits critical limitations when confronted with high-variance 
stochastic delays. The framework relies on a 
single-step state predictor that updates only upon arrival of a new 
delayed observation; between arrivals, the predictor output is held 
constant, and at each new arrival, the predicted state jumps 
discontinuously to incorporate the new measurement. These step-like 
discontinuities propagate into the derivative term of the downstream 
PD law, injecting spurious high-frequency content that destabilizes 
the control loop regardless of gain adaptation. Furthermore, PMDC's 
state prediction lacks explicit velocity learning, resulting in 
imprecise derivative estimates that compound with delay uncertainty 
under high-variance conditions. Finally, validation of these advanced predictive methods has been constrained primarily to simulated environments; their robustness and practical applicability on real-world robotic hardware under high-variance network conditions remain largely undemonstrated.

\subsection{Contributions}
To address these persisting research gaps, this paper proposes a hybrid framework that combines a delay-resilient state estimator with a residual reinforcement learning (DR-RL) policy for teleoperation under high-variance stochastic delays.
Our primary contributions are:
\begin{itemize} 
    \item Autoregressive State Estimation for Continuous Predictions: 
    We develop an LSTM-based state estimator that produces state estimates at a high control rate, decoupling the estimator output frequency from the stochastic and intermittent communication rate. 
    Unlike prior single-step predictors that update only upon packet arrival and produce discontinuous state estimates, our estimator combines a residual update head with an anchored autoregressive rollout, supplying the downstream control loop with kinematically continuous predictions while bounding the accumulated autoregressive drift to the inter-arrival interval of leader observations.

    \item Residual Reinforcement Learning for Robust Compensation: We propose a hybrid control architecture in which a residual RL agent learns corrective torque terms to compensate for model uncertainties and external disturbances. 
    The residual specifically compensates for two sources of controller mismatch observable during simulated training: (1) tracking deviations induced by imperfect predictions from the LSTM estimator under stochastic delay, and (2) residual errors that the underlying PD feedback gains alone are insufficient to suppress.

    \item Validation under Diverse Network Conditions: We validate our framework on a Franka Panda manipulator across three distinct delay scenarios: (1) low delay with low variance, (2) high delay with low variance, and (3) high delay with high variance. This comprehensive evaluation demonstrates practical robustness under varying stochastic communication conditions and addresses a critical gap in the experimental validation of delay-resilient teleoperation methods. 
\end{itemize}

\section{Preliminary and Problem Setting}

\subsection{Teleoperation System}
\label{subsec:system_baseline}
This work considers a homogeneous teleoperation system where the leader ($l$) and follower ($f$) robots share the same embodiment. 
We model each agent as a rigid-body mechanical system with $n$ degrees of freedom, whose robot motion equations are described by
\begin{equation}\label{eq_robot_dynamics}
\boldsymbol{M}(\boldsymbol{q}) \ddot{\boldsymbol{q}} + \boldsymbol{C}(\boldsymbol{q}, \dot{\boldsymbol{q}}) \dot{\boldsymbol{q}} + \boldsymbol{D}\dot{\boldsymbol{q}} + \boldsymbol{g}(\boldsymbol{q}) + \boldsymbol{d}(\boldsymbol{q}, \dot{\boldsymbol{q}}) = \boldsymbol{\tau},
\end{equation}
where the $\boldsymbol{q} \in \mathbb{R}^n$ is the joint positions, the matrices $\boldsymbol{M}(\boldsymbol{q})$, $\boldsymbol{C}(\boldsymbol{q}, \dot{\boldsymbol{q}})$, and $\boldsymbol{D} \in \mathbb{R}^{n \times n}$ are the inertia, Coriolis/centrifugal, and damping matrices, respectively. The vector function $ \boldsymbol{g}(\boldsymbol{q}) \in \mathbb{R}^n$ denotes the gravitational torques, and the control torque vector is $ \boldsymbol{\tau}\in \mathbb{R}^n$.The external disturbances and model uncertainties are considered in the model as a vector function $\boldsymbol{d}(\boldsymbol{q}, \dot{\boldsymbol{q}}) \in \mathbb{R}^n$.

\subsection{Control Objective}
\label{subsec:control_objective}
The primary control objective is to achieve high-fidelity motion synchronization between the leader and follower robots. Let $\boldsymbol{q}_l(t) \in \mathbb{R}^n$ and $\boldsymbol{q}_f(t) \in \mathbb{R}^n$ denote the vector of joint angular positions for the leader robot and the follower robot at time $t$, respectively. 
Since the human operator only controls the end-effector position, the control signal in Fig.~\ref{fig_Teleoperation_architecture} is considered as task-space motion. 
Here, we define the  \(\boldsymbol{x}_l(t) = \boldsymbol{f}_k(\boldsymbol{q}_l(t)) \in \mathbb{R}^m\) as the end-effector positions of the local robot, where \(\boldsymbol{f}_k: \mathbb{R}^n \to \mathbb{R}^m\) is the forward kinematics mapping. 

Therefore, we formally define the joint tracking error $\boldsymbol{e}(t)$ as the instantaneous deviation between the leader's and follower's configurations:
\begin{equation}
\label{eq_e}
    \boldsymbol{e}(t) = \boldsymbol{f}_k^{-1}(\boldsymbol{x}_l(t)) - \boldsymbol{q}_f(t).
\end{equation}
The goal is to design a control policy that minimizes the magnitude of the error $\boldsymbol{e}(t)$ under stochastic delays, including state delay $\omega_s \in \mathbb{R}_{>0}$, action delay $\omega_a \in \mathbb{R}_{>0}$, and observation delay $\omega_o \in \mathbb{R}_{>0}$.

\subsection{State Estimator}
\label{subsec:State Estimator}
Due to the inevitable communication delays between the leader and the follower, obtaining the real-time positions is infeasible. 
Consequently, the true leader state $\boldsymbol{q}_l(t), \dot{\boldsymbol{q}}_l(t)$ is unobservable at the current time step. The control agent only has access to the delayed states: $\boldsymbol{q}_l(t - \omega_s^t),~ \dot{\boldsymbol{q}}_l(t - \omega_s^t)$. 
However, using this delayed signal in a feedback loop would result in instability. 
To mitigate this latency and approximate the ground-truth state of the leader robot, we employ a LSTM-based state estimator to reconstruct the real-time leader robot state:  $\hat{\boldsymbol{q}}_l(t)$ and $\hat{\dot{\boldsymbol{q}}}_l(t)$, which denote the predicted leader states provided by the estimator in  Section \ref{subsec:lstm_estimator}.

\subsection{Controller Design}
To track the desired motion of the local leader, a nominal controller based on the computed torque control method is proposed as follows
\begin{subequations}
    \begin{align}
        \label{eq:tau}
    \boldsymbol{\tau}_{\mathrm{nom}} &= \boldsymbol{M}(\boldsymbol{q}_f) \ddot{\boldsymbol{q}}_{\mathrm{ref}} + \boldsymbol{C}(\boldsymbol{q}_f, \dot{\boldsymbol{q}}_f) \dot{\boldsymbol{q}}_f + \boldsymbol{D} \dot{\boldsymbol{q}}_f + \boldsymbol{g}(\boldsymbol{q}_f), \\
    \label{eq:q_ref}
\ddot{\boldsymbol{q}}_{\mathrm{ref}} &= \ddot{\hat{\boldsymbol{q}}}_l + \boldsymbol{K}_d \left( \hat{\dot{\boldsymbol{q}}}_l - \dot{\boldsymbol{q}}_f \right) + \boldsymbol{K}_p \left( \hat{\boldsymbol{q}}_l - \boldsymbol{q}_f \right),
    \end{align}
\end{subequations}
where $\boldsymbol{K}_p, \boldsymbol{K}_d \in \mathbb{R}^{n \times n}$ are the positive-definite proportional and derivative gain matrices.

Although the nominal controller provides a robust theoretical basis for trajectory tracking by compensating for known rigid-body dynamics, its efficacy is inherently constrained by unmodeled dynamics and state estimation errors. 
Inspired by \citep{Johannink_ICRA_2019_Residual}, we train a reinforcement learning agent to learn the residual torque compensation term $\boldsymbol{\tau}_{\mathrm{rl}}$. 
Therefore, the final control input $\boldsymbol{\tau}_{\mathrm{cmd}}$ applied to the follower robot is formulated as the summation of the nominal baseline and the learned residual, i.e.,
\begin{equation}
    \label{eq:hybrid_control_intro}
    \boldsymbol{\tau}_{\mathrm{cmd}} = \boldsymbol{\tau}_{\mathrm{nom}} + \boldsymbol{\tau}_{\mathrm{rl}},
\end{equation}
where the $\boldsymbol{\tau}_{\mathrm{rl}}$ is introduced in Section~\ref{subsec:residual_rl}.

\section{Methodology}
\label{sec:methodology}

\subsection{LSTM-based State Estimator}
\label{subsec:lstm_estimator}
A critical issue in teleoperation under stochastic delays is the 
discrete jumps in the feedback signal from the local robot. 
Such discontinuities result in unbounded derivative 
estimates in the downstream control loop, regardless of the specific 
controller used, and induce high-frequency oscillations and chattering. 
Thus, the estimator must guarantee smooth trajectory reconstruction in addition to minimizing state error, so that the controller can operate on a continuous input signal.

\subsubsection{Residual Update Head}
\label{subsec:residual_update_head}
Let the set $\mathcal{Z}(t)$ denote the sequence of the $N^t$ most recent 
delayed states available at time $t$:
\begin{equation}
    \mathcal{Z}(t) = \{ \mathbf{s}_{i}^t\}_{i=1,\dots, N^t},
\end{equation}
where the element $\mathbf{s}_i^t = [\boldsymbol{q}_l(t)^{\top}\!, 
{\dot{\boldsymbol{q}}_l(t)}^{\top}\!, \omega_s^t]^{\top}$ represents 
the concatenation of the joint positions $\boldsymbol{q}_l(t)$, 
velocities ${\dot{\boldsymbol{q}}_l(t)}$, and the state delay magnitude 
$\omega_s^t$.
The LSTM processes this input sequence and maintains an internal 
recurrent state $(\mathbf{h}_k, \mathbf{c}_k)$ that summarizes the 
recent leader motion. The hidden output $\mathbf{h}_k$ is consumed by 
the feed-forward head defined below.
To design this head, we first observe that the underlying robot 
dynamics admit an accumulative structure: the transition from state 
$\mathbf{s}_{k-1}$ to $\mathbf{s}_k$ over one integration step is 
governed by a nonlinear function $\ell$ representing the rigid-body 
dynamics and integration scheme,
\begin{equation}
    \label{eq:general_dynamics}
    \mathbf{s}_k^t = \mathbf{s}_{k-1}^t + \ell(\mathbf{s}_{k-1}^t)\, \Delta t \approx \mathbf{s}_{k-1}^t + \Delta \mathbf{s}_{\mathrm{dyn}},
\end{equation}
where $\Delta \mathbf{s}_{\mathrm{dyn}}$ denotes the per-step state change induced 
by velocity, acceleration, Coriolis forces, and gravity.
To mirror this accumulative structure inside the network, we do not let the LSTM regress the next absolute state directly. Instead, a small feed-forward head maps the LSTM hidden representation $\mathbf{h}_k$ to a learned increment $\Delta \mathbf{s}_k$, which is composed with the previous state through an explicit Euler update:
\begin{equation}
    \label{eq:euler_update}
    \hat{\mathbf{s}}_k = \mathbf{s}_{k-1} + \Delta \mathbf{s}_k, \qquad \Delta \mathbf{s}_k = \mathbf{v}_k \cdot \alpha,
\end{equation}
where $\mathbf{v}_k$ is the network-predicted velocity correction and 
$\alpha \in \mathbb{R}_{>0}$ is a learnable scaling factor. 
This residual formulation aligns the network output structurally with 
the rigid-body increment $\Delta \mathbf{s}_{\mathrm{dyn}}$ in 
Eq.~\eqref{eq:general_dynamics}, keeps the regression target 
well-conditioned over the small per-step magnitude, and guarantees 
kinematic continuity between consecutive predictions by 
construction rather than as a soft regularizer in the loss function.

\subsubsection{Autoregressive State Estimation}
To generate continuous state estimates at the control rate over 
extended delay periods, we employ an autoregressive prediction 
strategy. This mechanism leverages the temporal memory in the LSTM 
hidden state together with the residual update in 
Eq.~\eqref{eq:euler_update} to simulate the robot's forward motion 
from the most recent anchor observation $\hat{\mathbf{s}}_k^{(t-\omega_s^t)}$ 
up to the current time $t$. The prediction horizon is determined by 
the maximum observed system delay $\omega_s^{\max}$. During the delay 
interval, the estimator performs autoregressive rollouts at the 
control rate, where the model uses its own previous output as input 
for the subsequent prediction step:
\begin{equation}
\hat{\mathbf{s}}_{k}^{(t-\omega_s^t)} = \hat{\mathbf{s}}_{k-1}^{(t-\omega_s^t)} + \Delta \mathbf{s}_k,
\end{equation}
where $\Delta \mathbf{s}_k$ is the increment produced by the LSTM 
head conditioned on the predicted state from the previous iteration 
and the running hidden state. The autoregressive 
chain is anchored to the most recent delayed leader observation 
rather than maintained continuously across the trajectory. Whenever 
a new delayed observation $\mathbf{s}^{(t-\omega_s^t)}$ arrives at 
the agent, the previous autoregressive predictions are discarded 
and the rollout is re-initialized from this anchor, with the LSTM  hidden state correspondingly reset to the value produced by re-encoding the input window terminating at the new anchor.

\subsection{Residual RL Control}
\label{subsec:residual_rl}

\subsubsection{Observation Space}
The presence of stochastic time-varying delay $\omega_s^t$ inherently renders the environment partially observable, as the agent cannot access the immediate true state of the leader robot. This effectively transforms the standard MDP into a POMDP. A naive approach is State Augmentation (stacking history states into the agent observation space) \citep{Nath_ACM2021_RevisitingStateAugmentation}, however, this leads to the curse of dimensionality as the required augmentation history length $N$ grows with delay magnitude. 
To avoid this computational intractability, we adopt a Model-Based Reinforcement Learning approach. We rely on the LSTM estimator (Section \ref{subsec:lstm_estimator}) to estimate real-time states $\hat{\boldsymbol{q}}_l(t)$ and $\hat{\dot{\boldsymbol{q}}}_l(t)$. This effectively transforms the POMDP into a standard MDP process for the control policy.
The RL agent receives an augmented observation vector $\boldsymbol{o}_t \in \mathbb{R}^{d_{obs}}$ comprising the follower robot state  $[\boldsymbol{q}_f, \dot{\boldsymbol{q}}_f]^T$, the LSTM-predicted leader state $[\hat{\boldsymbol{q}}_l, \hat{\dot{\boldsymbol{q}}}_l]^T$, the tracking errors $[\boldsymbol{e}_q, \boldsymbol{e}_{\dot{q}}]^T$, and the follower state history $\{ \boldsymbol{q}_f(t-k), \dot{\boldsymbol{q}}_f(t-k) \}_{k=1}^{N}$.

\subsubsection{SAC Algorithm}
In this work, we employ Soft Actor-Critic (SAC) for policy optimization due to its sample efficiency and stability in continuous control tasks. 

\textbf{Objective.}
The standard reinforcement learning objective is defined as the 
expectation of the discounted cumulative reward 
$R(\tau) = \sum_{t=0}^{T} \gamma^t r(\mathbf{s}_t, \mathbf{a}_t)$ 
over trajectories sampled under the policy $\pi_\theta$:
\begin{equation}
    J(\theta) = \mathbb{E}_{\tau \sim P_{\pi_{\theta}}} [R(\tau)].
\end{equation}

\textbf{Value functions.}
To evaluate the policy, SAC relies on the standard state-value and 
action-value functions:
\begin{align}
    V^{\pi}(\mathbf{s}) &= \mathbb{E}_{\pi} \left[ \sum_{k=0}^{\infty} \gamma^k r_{t+k} \,\middle|\, \mathbf{s}_t = \mathbf{s} \right], \\
    Q^{\pi}(\mathbf{s}, \mathbf{a}) &= \mathbb{E}_{\pi} \left[ \sum_{k=0}^{\infty} \gamma^k r_{t+k} \,\middle|\, \mathbf{s}_t = \mathbf{s}, \mathbf{a}_t = \mathbf{a} \right].
\end{align}
The action-value function $Q^\pi$ serves as the Critic, providing 
the gradient signal used to improve the actor.

\textbf{Entropy-regularized objective.}
SAC augments the standard objective with a per-step entropy term 
$\mathcal{H}(\pi(\cdot|\mathbf{s}_t))$:
\begin{equation}
    J_{\text{SAC}}(\theta) = \mathbb{E}_{\tau \sim \pi_\theta} \left[ \sum_{t=0}^{T} \gamma^t \left( r(\mathbf{s}_t, \mathbf{a}_t) + \alpha\, \mathcal{H}(\pi(\cdot|\mathbf{s}_t)) \right) \right],\!\!
\end{equation}
where $\alpha$ is the temperature parameter governing the trade-off 
between reward maximization and entropy preservation. Maximizing 
entropy encourages the agent to explore multiple modes of high-reward 
behavior rather than collapsing into a single deterministic policy, 
which is particularly important under stochastic delays where the 
optimal action distribution may be inherently multi-modal.

\textbf{Soft Bellman backup.}
Under the entropy-regularized objective, the Critic is trained to 
satisfy the soft Bellman equation, which incorporates the entropy 
bonus directly into the value backup:
\begin{equation}
    Q^\pi(\mathbf{s}, \mathbf{a}) = r + \gamma\, \mathbb{E}_{\mathbf{s}', \mathbf{a}' \sim \pi}\!\left[ Q^\pi(\mathbf{s}', \mathbf{a}') - \alpha \log \pi(\mathbf{a}'|\mathbf{s}') \right],
\end{equation}
where the prime denotes the next time step. Compared to the standard 
Bellman backup, the additional term $-\alpha \log \pi(\mathbf{a}'|\mathbf{s}')$ 
assigns higher value to states from which the future policy can act 
with greater entropy.

\textbf{Policy improvement.}
Given the updated Critic, the actor is improved by projecting the 
policy onto the distribution induced by the exponentiated soft 
Q-function:
\begin{equation}
    \pi_{\text{new}} = \arg\min_{\pi' \in \Pi}\, D_{\mathrm{KL}}\!\left( \pi'(\cdot|\mathbf{s})\, \middle\|\, \frac{\exp(Q^\pi(\mathbf{s}, \cdot) / \alpha)}{Z^\pi(\mathbf{s})} \right),
\end{equation}
where $Z^\pi(\mathbf{s})$ is the normalization constant. This update 
preserves entropy in the resulting policy while pulling probability 
mass towards high-Q actions, yielding the alternating Critic-Actor 
optimization that defines SAC.

\subsubsection{Multi-Objective Reward Design}
\label{subsec:reward}
The control policy optimization is driven by a composite reward 
function $r: \mathcal{S} \times \mathcal{A} \rightarrow \mathbb{R}$, 
formulated to balance asymptotic tracking stability with control 
effort minimization and strict safety constraints. We employ a 
linear scalarization of these objectives, defined at each time 
step $t$ as:
\begin{equation}
    r_t(\mathbf{s}_t, \mathbf{a}_t) = r_{\text{track}} + r_{\text{reg}} + r_{\text{safe}}.
\end{equation}

The {tracking fidelity} term $r_{\text{track}}$ penalizes the 
weighted squared Euclidean distance between the follower joint state 
and the LSTM-estimated leader joint state:
\begin{equation}
    r_{\text{track}} = - \lambda_p \| \hat{\mathbf{q}}_l - \mathbf{q}_f \|_2^2 - \lambda_v \| \hat{\dot{\mathbf{q}}}_l - \dot{\mathbf{q}}_f \|_2^2,
\end{equation}
where $\lambda_p, \lambda_v \in \mathbb{R}_{>0}$ are weighting 
coefficients governing the trade-off between position accuracy and 
velocity synchronization.
The {action regularization} term {$r_{\text{reg}}$} 
discourages large residual torques:
\begin{equation}
    r_{\text{reg}} = - \lambda_a \frac{1}{n} \sum_{i=1}^{n} (a_{t,i})^2,
\end{equation}
where $a_{t,i}$ is the residual torque applied to joint $i$.
The {safety penalty} term $r_{\text{safe}}$ enforces hard 
constraints on the agent's behavior as the sum of three per-step 
penalty terms:
\begin{equation}
    r_{\text{safe}} = P_{\text{joint}} + P_{\text{track}} + P_{\text{est}},
\end{equation}
where each penalty is independently activated upon violation of its 
corresponding condition:
\begin{equation}
\begin{cases}
 P_{\text{joint}} = \rho, & \text{if } \exists i,~ q_{f,i} \notin [q_{\min}^i, q_{\max}^i], \\
 P_{\text{track}} = \rho, & \text{if } \| \boldsymbol{e}_{\text{track}} \| > \epsilon_{\text{track}}, \\
 P_{\text{est}} = \rho, & \text{if } \| \boldsymbol{e}_{\text{est}} \| > \epsilon_{\text{est}}, \\
\end{cases},
\end{equation}
where $\rho < 0$ is a fixed per-step penalty magnitude, and the 
three terms accumulate when multiple conditions are simultaneously 
violated. The $[q_{\min}^i, q_{\max}^i]$ denotes the physical limits 
of the $i$-th joint. The tracking error and estimation error, 
computed during training using the ground-truth leader state 
available in simulation, are defined as:
\begin{equation}
    e_{\text{track}}(k) = \| \boldsymbol{q}_l(k) - \boldsymbol{q}_f(k) \|_2, \quad
    e_{\text{est}}(k) = \| \hat{\boldsymbol{q}}_l(k) - \boldsymbol{q}_l(k) \|_2.
\end{equation}
To stabilize critic learning under the per-step safety penalties, the composite reward $r_t$ is clipped to a bounded range $[r_{\min}, r_{\max}]$ during training\footnote{The detailed training hyperparameters are listed in Appendix~\ref{subsec:training}}. 

\section{Experimental Results}
\subsection{Experimental Setup}
\label{subsec:exp_setup}
Our experimental evaluation consists of two parts. 
First, we conduct a systematic quantitative comparison between DR-RL 
and two baseline methods in a MuJoCo simulation of the Franka Panda manipulator, evaluated under 
three stochastic delay configurations (Section~\ref{subsec:result_analysis}). 
Second, we deploy the simulator-trained DR-RL policy on the physical Franka Panda manipulator to 
verify that the framework transfers to real hardware without 
retraining (Section~\ref{subsec:real_deployment}).

We systematically compare our method against two baseline approaches:
\begin{itemize}
\item \textbf{PMDC (SOTA model-based RL method)}~\citep{McCutcheon_IROS2023_Adaptive}: 
Utilizes a learned dynamics model and State-Buffer based State 
Prediction (SBSP) to explicitly estimate the real-time state from 
delayed observations.

\item \textbf{Vanilla PD (Baseline method)}: A standard inverse 
dynamics controller operating directly on delayed observations 
without prediction or compensation, serving as the performance 
lower bound.
\end{itemize}

To quantify the tracking performance, a desired reference trajectory 
for the end-effector is commanded to the leader robot, which the 
follower must replicate under delay. The reference trajectory in 
the task space is defined as
\begin{align}
    \boldsymbol{x}_l^{\mathrm{ref}}(k) = [
0.2\sin(3k),
0.2\sin(4k + \frac{\pi}{2}),
0.02\sin(k)]^{\top}.
\end{align}
The teleoperation system architecture deploys the RL agent co-located 
with the follower robot, resulting in constant action delay 
$\omega_o^t = \omega_a^t = 50$ ms between the agent's torque commands 
and the follower robot's execution for all $t$. Stochasticity arises 
from the network communication between the leader robot and the RL 
agent, introducing variable observation delay 
$\omega_s^t \sim \mathcal{U}(\omega_{\min}, \omega_{\max})$. The 
sampling period is set as $\Delta t = 4$ ms.
We evaluate three delay configurations as follows:
\begin{itemize}
    \item \textbf{Low delay with low variance}: $\omega_s^t \sim \mathcal{U}(120, 160)$ ms, total: 170--210 ms
    \item \textbf{High delay with low variance}: $\omega_s^t \sim \mathcal{U}(200, 240)$ ms, total: 250--290 ms
    \item \textbf{High delay with high variance}: $\omega_s^t \sim \mathcal{U}(40, 240)$ ms, total: 90--290 ms
\end{itemize}
These three delay configurations are used both in the simulation comparison and in the physical-robot deployment of DR-RL, allowing a consistent evaluation of the framework across the operating envelope of stochastic delays considered in this work.

\subsection{Quantitative Comparison in Simulation}
\label{subsec:result_analysis}
We evaluate the tracking performance by the norm of the Cartesian tracking error $\|\boldsymbol{e}(t)\|$ between the leader and follower end-effectors, computed over the first 50 s of each trial\footnote{The additional results are available in Appendix~\ref{app_ar}.}. 
In Fig.~\ref{fig_drt_pd_pmdc_tracking_error_comparison}, it is shown that in all three scenarios, DR-RL is superior to PD and PMDC methods.
Under low-delay, low-variance conditions (170--210 ms), DRT-RL achieves the lowest tracking error, outperforming both PMDC and the Vanilla PD controller, which exhibits substantial oscillations (peaks $\approx$ 0.4 m); this demonstrates the superiority of LSTM-based autoregressive estimation over feedforward networks used in PMDC. 
As delay magnitude increases to 250--290 ms, DRT-RL maintains stable, low-error performance, whereas PMDC deteriorates and the PD controller fails completely with error peaks exceeding 1 m. 
The most pronounced differentiation emerges under the challenging high-delay, high-variance scenario (90--290 ms): the PD controller exhibits severe instability with consistent oscillation, and PMDC degrades significantly due to the inability of its memoryless feedforward network to adapt to variable delays from fixed-length inputs.

\begin{figure}[t]
    \centering
    \includegraphics[width=1\linewidth]{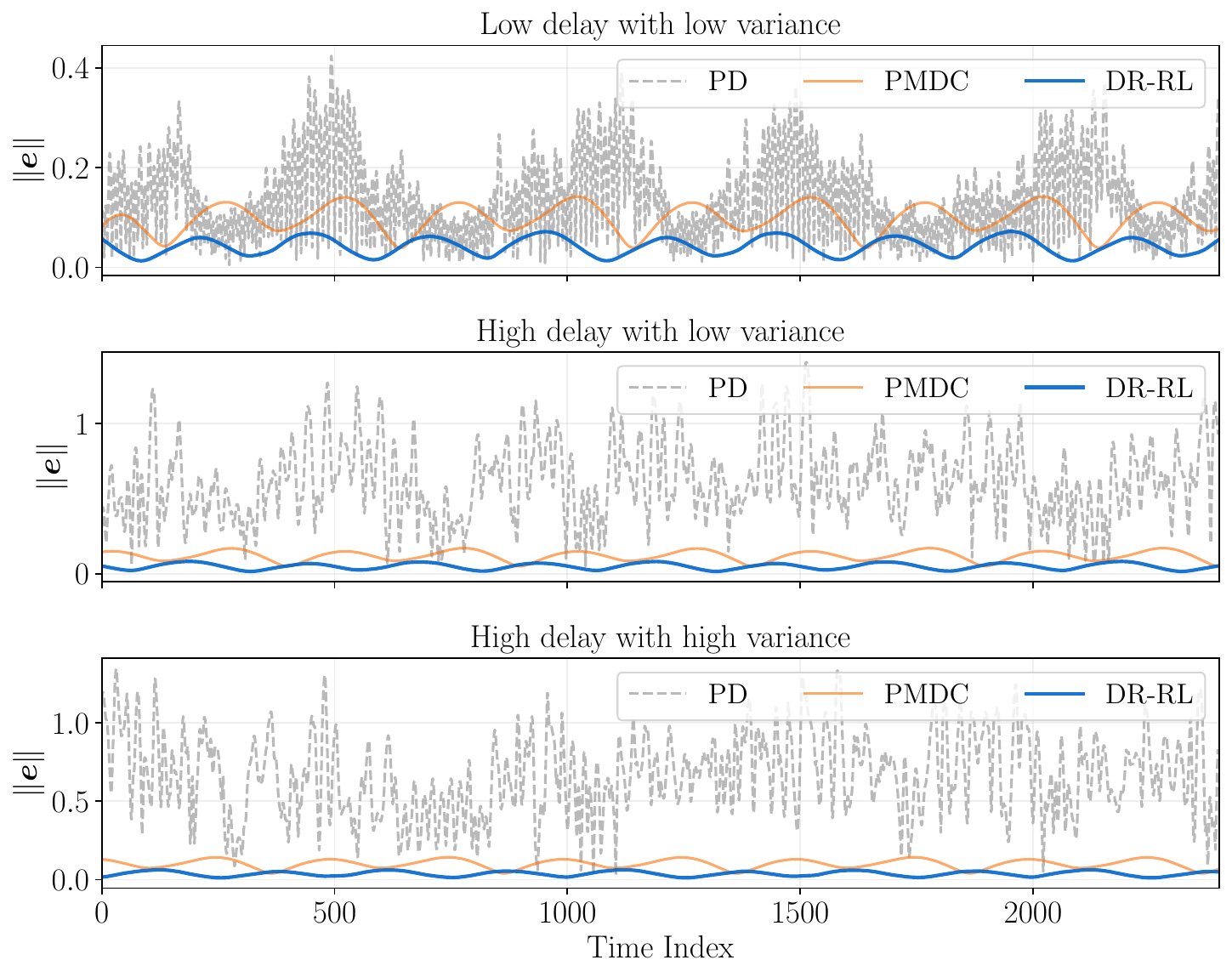}
    \caption{Tracking error $\|\boldsymbol{e}\|$ comparison against 
    other methods under different network delay conditions.}
    \label{fig_drt_pd_pmdc_tracking_error_comparison}
\end{figure}

\subsection{Real-Robot Deployment}
\label{subsec:real_deployment}
To verify that the simulator-trained DR-RL policy 
transfers to physical hardware, we deploy the same policy and 
estimator weights used in the simulation study on a physical Franka 
Panda manipulator, with no additional fine-tuning or calibration on 
real-robot data. Table~\ref{tab:sim2real} reports the resulting 
sim-to-real tracking error under the high-delay, high-variance 
configuration ($90\text{--}290$~ms), which corresponds to the 
hardest of the three settings evaluated in the simulation.
\begin{table}[!t]
\centering
\caption{Sim-to-real tracking error comparison of DR-RL under the high-delay, high-variance delays.}
\label{tab:sim2real}
\small
\setlength{\tabcolsep}{6pt}
\begin{tabular}{lccc}
\toprule
 & Sim & Real & Ratio \\
\midrule
Mean (m) & 0.037 & 0.045 & 1.22$\times$ \\
Max (m)  & 0.063 & 0.078 & 1.24$\times$ \\
\bottomrule
\end{tabular}
\end{table}

\begin{figure}[t]
    \centering
    \subfloat[MuJoCo environment.]{
        \includegraphics[width=0.49\linewidth]{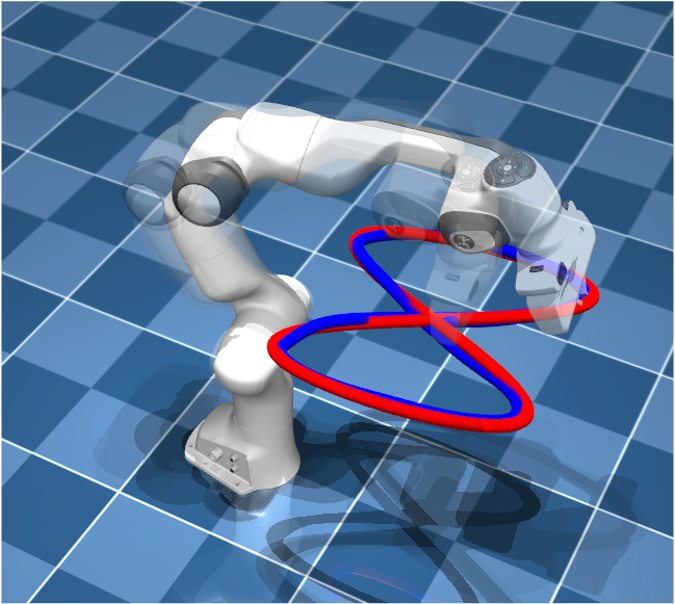}
    }
    \subfloat[Real experiment.]{
        \includegraphics[width=0.5\linewidth]{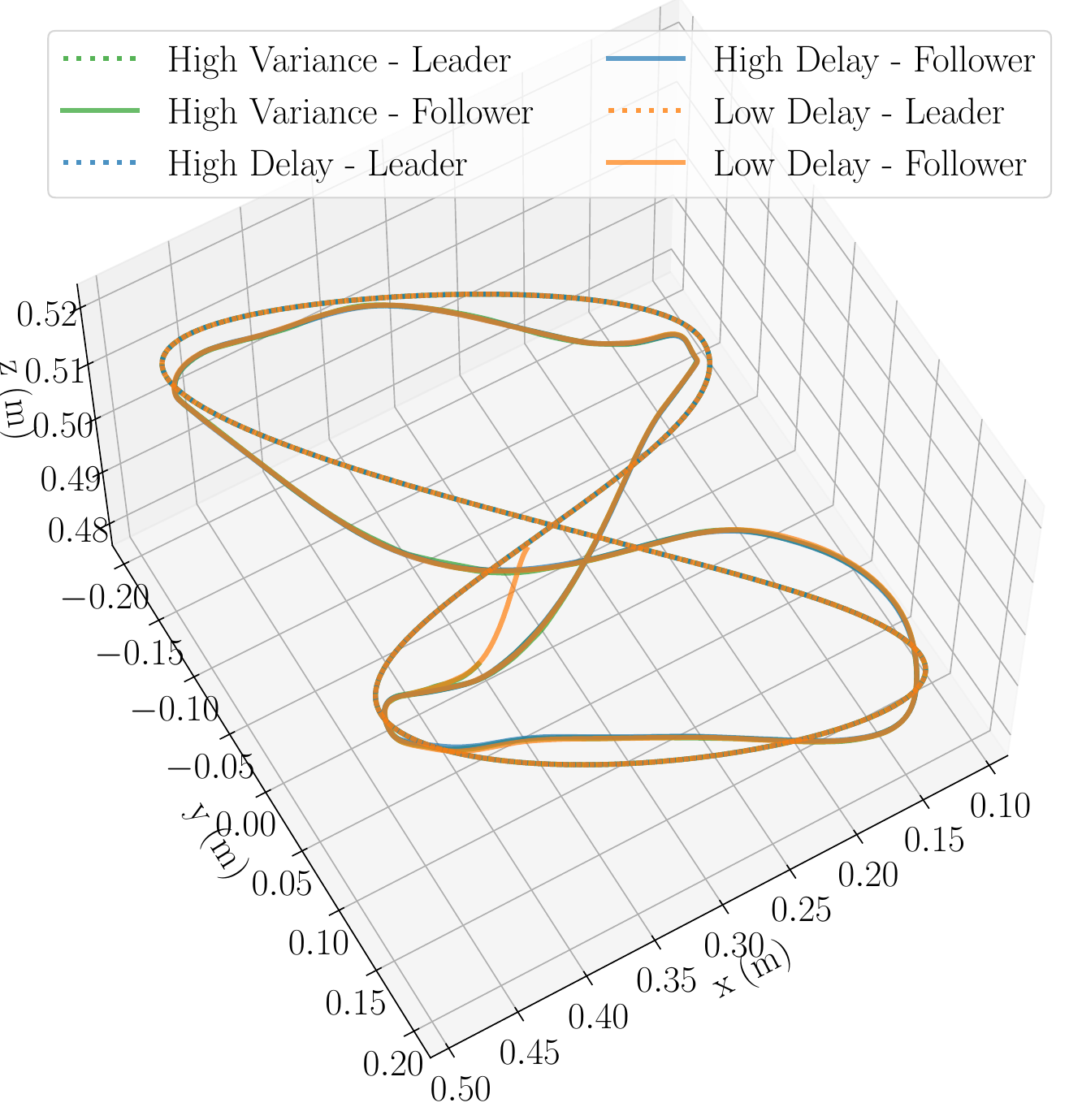}
    }
    \caption{Visualization of spatial trajectories using DR-RL in 
    a high-delay, high-variance scenario. (a) The red line is the 
    local leader, the blue line denotes the remote follower. 
    (b) The dashed line is the leader, while the solid line is the 
    follower.}
    \label{fig_trajectories}
\end{figure}

The mean Cartesian tracking error increases from 
$0.037$~m in simulation to $0.045$~m on hardware, a relative 
degradation of approximately $22\%$. The maximum error rises by a 
similar factor. This bounded sim-to-real gap reflects three 
properties of the proposed framework. First, the LSTM estimator 
operates only on leader-side kinematic trajectories, which are 
independent of follower-robot dynamics and therefore consistent 
between simulation and physical deployment. Second, the nominal 
computed-torque controller in Eq.~\eqref{eq:tau} is parameterized 
with manufacturer-calibrated dynamics that are also used directly 
in the MuJoCo simulator, eliminating the inertial mismatch that 
typically dominates sim-to-real degradation in residual learning 
frameworks. Third, the residual RL policy learns only a small 
torque correction on top of the nominal controller, bounding the 
magnitude by which sim-trained behavior can degrade under unmodeled 
real-world dynamics. The corresponding spatial trajectory on the 
physical manipulator is shown in Fig.~\ref{fig_trajectories}b.

\section{Conclusion}
This paper introduces DR-RL, a hybrid learning-based 
control framework for robust teleoperation under high-variance 
stochastic delays. 
At its core is an autoregressive state estimator that leverages LSTM-based temporal memory to resolve delay-induced partial observability, while a coupled residual reinforcement learning policy compensates for unmodeled dynamics to ensure stable and precise operation.
Experimental validation on a physical Franka Panda demonstrates that our framework significantly outperforms SOTA baselines, achieving superior tracking performance and stability.

\bibliography{ref} 

\newpage
\appendix
\section{Training Procedure}
\label{subsec:training}
The proposed framework is trained in two stages, both conducted entirely within a MuJoCo simulation of the Franka Panda manipulator. No real-robot data is used during training; the physical experiments described in Section~\ref{subsec:result_analysis} therefore also serve as a sim-to-real evaluation. Training and validation environments are instantiated from distinct random seeds to prevent data leakage.

\textbf{Training data and collection.}
The training data consists of figure-8 reference trajectories generated on-the-fly by the leader simulator, with their geometric and temporal parameters randomized at each episode reset (center $c_x \in [0.3, 0.4]$~m, $c_y \in [-0.1, 0.1]$~m, scale $s_{x,y} \in [0.1, 0.3]$~m, $s_z \in [0.01, 0.03]$~m, frequency $f \in [0.05, 0.15]$~Hz). At every control step, the leader joint state, the corresponding stochastic delay $\omega_s^t$, and the future ground-truth trajectory used as the autoregressive target are written into a circular replay buffer.
This online collection scheme is chosen over a pre-collected fixed dataset for two reasons. First, the joint distribution of trajectory shape and delay realization is too high-dimensional to enumerate offline; sampling on-the-fly ensures uniform coverage of the operating envelope used at deployment. Second, the autoregressive target tensor (245 future steps per sample) makes a fully materialized dataset prohibitively large; the buffer keeps only the most recent samples and overwrites older ones, providing a near-uniform on-policy distribution at constant memory cost.

\textbf{Estimator training.}
In the first stage, the LSTM estimator is trained to minimize the mean squared error between its autoregressive rollout and the ground-truth trajectory over the AR horizon. A naive autoregressive loss diverges in practice because the network never receives its own (initially poor) predictions during training; we therefore employ scheduled sampling, starting from full teacher forcing and decaying the ground-truth probability geometrically per update. This gradually exposes the network to its own error distribution and stabilizes long-horizon rollouts. Early stopping is governed by a closed-loop validation rollout in which the estimator is plugged into the simulation and the resulting prediction error is monitored, so that the validation metric reflects deployment-time performance rather than open-loop one-step accuracy.

\textbf{Policy training.}
In the second stage, the residual SAC policy is trained with the LSTM estimator weights frozen, so that the policy treats the estimator as a fixed component of the environment. Predicted leader states are supplied as part of the observation vector defined in Section~\ref{subsec:residual_rl}. Transitions collected during the warm-up phase are excluded from the replay buffer because the leader state history during this period is incomplete and would bias the value estimates. The best-performing checkpoint is retained based on the mean return of deterministic validation rollouts. Key hyperparameters of both stages are summarized in Table~\ref{tab:training_hyperparams}.

\begin{table}[!t]
\centering
\caption{Key training hyperparameters of the LSTM estimator and the residual SAC policy.}
\label{tab:training_hyperparams}
\small
\setlength{\tabcolsep}{4pt}
\begin{tabular}{ll}
\toprule
\textbf{Parameter} & \textbf{Value} \\
\midrule
\multicolumn{2}{l}{ {\textit{LSTM estimator}}} \\
 {Hidden dim.\ / layers} &  {256 / 3} \\
 {Output head activation} &  {Mish} \\
 {Input seq.\ length} &  {150 steps (600 ms)} \\
 {AR rollout horizon} &  {245 steps ($\approx 980$ ms)} \\
 {Learning rate / batch size} &  {$5 \times 10^{-4}$ / 256} \\
 {Total updates} &  {500{,}000} \\
 {Scheduled sampling decay} &  {0.99995 per update} \\
\midrule
\multicolumn{2}{l}{ {\textit{Residual SAC policy}}} \\
 {Hidden dims (actor / critic)} &  {[512, 256]} \\
 {LR (actor / critic / $\alpha$)} &  {$3{\times}10^{-5}$ / $3{\times}10^{-5}$ / $3{\times}10^{-4}$} \\
 {Discount $\gamma$ / soft update $\tau$} &  {0.99 / 0.005} \\
 {Batch size / buffer size} &  {128 / $1 \times 10^{6}$} \\
 {Total environment steps} &  {$3 \times 10^{6}$} \\
 {Reward weights ($\lambda_p, \lambda_v, \lambda_a$)} &  {10, 5, 0.01} \\
 {Safety thresholds ($\epsilon_{\text{track}}, \epsilon_{\text{est}}$)} &  {0.5, 0.3 rad} \\
 {Safety penalty $\rho$} &  {$-5$} \\
 {Reward clip range ($r_{\min}, r_{\max}$)} &  {$[-20, 5]$} \\
\midrule
\multicolumn{2}{l}{ {\textit{Common}}} \\
 {Optimizer} &  {Adam} \\
 {Control frequency} &  {250 Hz ($\Delta t = 4$ ms)} \\
\bottomrule
\end{tabular}
\end{table}

\section{Additional Results}
\label{app_ar}
\begin{figure}[!htbp]
    \centering
    \includegraphics[width=0.88\linewidth]{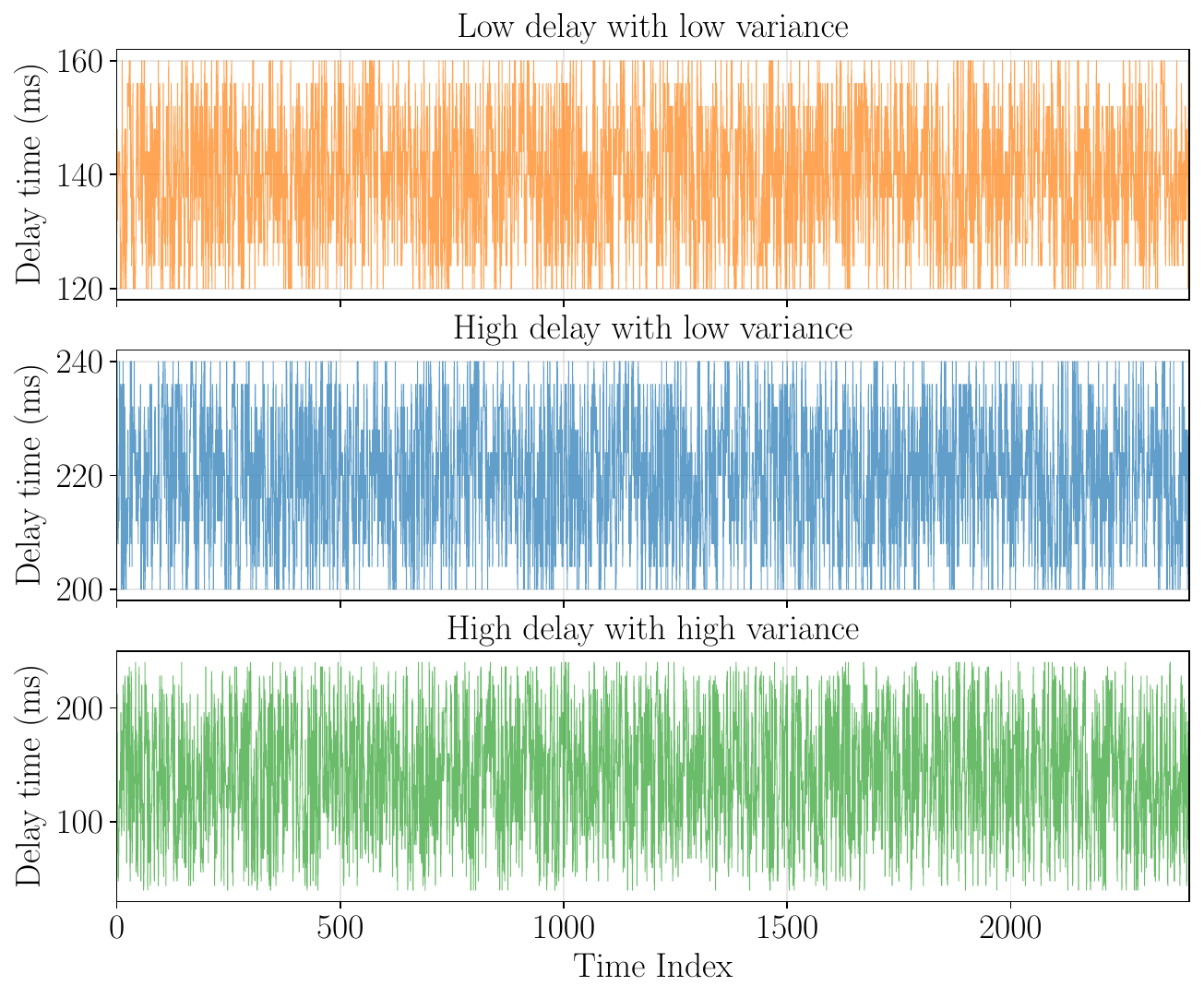}
    \caption{State delay patterns in teleoperation tasks. Low delay 
    with low variance (top), high delay with low variance (middle), 
    high delay with high variance (bottom).}
    \label{fig_drt_delay_plot}
\end{figure}

To visualize the three delay configurations in Fig.~\ref{fig_drt_delay_plot}.

Table~\ref{tab:sim_metrics} summarizes the mean, 95th percentile (P95), maximum, and standard deviation of $\|\boldsymbol{e}(t)\|$ 
for DR-RL, PMDC, and the Vanilla PD baseline across all three delay configurations.

Across all three configurations, DR-RL achieves the 
lowest mean tracking error, reducing the mean by approximately 
$58\text{--}64\%$ relative to the SOTA model-based RL baseline (PMDC). 
The improvement is consistent across percentiles: the 95th-percentile 
error of DR-RL is below the {mean} error of PMDC in every 
configuration, indicating that DR-RL's worst-case behavior 
out-performs PMDC's average-case behavior under the same delay 
conditions. Under low delay with low variance ($170\text{--}210$~ms), DR-RL attains a mean error of $0.036$~m versus $0.099$~m for PMDC in . 
As the delay magnitude increases to $250\text{--}290$~ms 
(high / low variance), the Vanilla PD baseline fails completely with 
maximum errors exceeding $1.4$~m, while DR-RL remains stable at 
$0.050$~m mean error. Under the most challenging configuration of 
$90\text{--}290$~ms (high / high variance), PD continues to exhibit 
severe oscillations and PMDC degrades to $0.100$~m mean error, 
whereas DR-RL maintains $0.037$~m. This robustness under 
high-variance delay reflects the combined effect of the autoregressive 
LSTM estimator, which supplies continuous predictions at the control 
rate, and the torque-level residual policy, which compensates for 
the residual model uncertainty rather than relying on gain adaptation.

\begin{table}[t]
\centering
\caption{Tracking error metrics in simulation.}
\label{tab:sim_metrics}
\small
\renewcommand{\arraystretch}{1.2}
\setlength{\tabcolsep}{4pt}
\begin{tabular}{lccc}
\toprule
Condition / Method & Mean (m) & P95 (m) & Max (m) \\
\midrule
\multicolumn{4}{l}{\textit{Low / low var.}} \\
\quad DR-RL & \textbf{0.036} & \textbf{0.059} & \textbf{0.062} \\
\quad PMDC   & 0.099 & 0.139 & 0.142 \\
\quad PD     & 0.24 & 0.371 & 0.509 \\
\midrule
\multicolumn{4}{l}{\textit{High / low var.}} \\
\quad DR-RL & \textbf{0.050} & \textbf{0.079} & \textbf{0.083} \\
\quad PMDC   & 0.119 & 0.168 & 0.172 \\
\quad PD     & 0.618 & 1.091 & 1.409 \\
\midrule
\multicolumn{4}{l}{\textit{High / high var.}} \\
\quad DR-RL & \textbf{0.037} & \textbf{0.060} & \textbf{0.063} \\
\quad PMDC   & 0.100 & 0.140 & 0.142 \\
\quad PD     & 0.671 & 1.098 & 1.346 \\
\bottomrule
\end{tabular}
\end{table}

\end{document}